# Evaluating the Readability of Force Directed Graph Layouts: A Deep Learning Approach

**Hammad Haleem**
The Hong Kong University of Science and Technology, Hong Kong, China

**Yong Wang**
The Hong Kong University of Science and Technology, Hong Kong, China

**Abishek Puri**
The Hong Kong University of Science and Technology, Hong Kong, China

**Sahil Wadhwa**
Blackrock, New Delhi, India

**Huamin Qu**
The Hong Kong University of Science and Technology, Hong Kong, China

Existing graph layout algorithms are usually not able to optimize all the aesthetic properties desired in a graph layout. To evaluate how well the desired visual features are reflected in a graph layout, many readability metrics have been proposed in the past decades. However, the calculation of these readability metrics often requires access to the node and edge coordinates and is usually computationally inefficient, especially for dense graphs. Importantly, when the node and edge coordinates are not accessible, it becomes impossible to evaluate the graph layouts quantitatively. In this paper, we present a novel deep learning-based approach to evaluate the readability of graph layouts by directly using graph images. A convolutional neural network architecture is proposed and trained on a benchmark dataset of graph images, which is composed of synthetically-generated graphs and graphs created by sampling from real large networks. Multiple representative readability metrics (including edge crossing, node spread, and group overlap) are considered in the proposed approach. We quantitatively compare our approach to traditional methods and qualitatively evaluate our approach by showing usage scenarios and visualizing convolutional layers. This work is a first step towards using deep learning based methods to quantitatively evaluate images from the visualization field.

## INTRODUCTION

Graphs have been widely used to represent network data from a variety of domains like social networks, biological networks, mobile device connection networks, financial transaction networks, etc. Graphs are often visualized as node-link diagrams, so as to support a variety of network exploration tasks. Accordingly, many automated graph layout techniques have been proposed in the past few decades,[5] which attempt to optimize various visual properties (readability metrics) for graph visualization.[3,19] For example, the traditional force-directed graph layout algorithm builds off a force model to avoid node occlusions and edge crossings,[4] and clustering-based graph layout techniques are designed to preserve the cluster structures of nodes.[15] There are also other graph layout methods that aim to preserve the relative graph-theoretic distances of nodes through dimensional reduction techniques. However, it is impossible to optimize all these desired properties in a graph layout, as the optimization goals of some desired properties essentially contradict each other. It means when researchers want to check whether a specific graph layout is desirable, they must evaluate the graph layout from different perspectives.

Prior work has introduced various quantitative measures, i.e. readability metrics,[9,19] to evaluate the graph layouts comprehensively. The most popular readability metrics for a graph layout include edge crossing, node overlapping, and others, measuring whether the graph is desirable or not from a certain perspective. More specifically, a high number of edge crossings affect the interpretation of connections between nodes, a large amount of node overlapping hinders the understanding of the node information accurately, and a significant number of overlapping node clusters make it difficult to identify the node groups. To quantitatively evaluate the readability of a graph layout, the general approach is to calculate the readability metrics by directly using the coordinates of nodes and edges. Two issues exist in this process. First, apart from rendering the graph visualization result in the 2D plane, we also need to store the positions of all the nodes and edges to prepare for the subsequent readability calculation. Secondly, the calculation of readability metrics using the coordinates of nodes and edges is usually not efficient. For example, for a graph $G = (V, E)$, the calculation of node overlapping and edge crossing through pairwise comparison has a time complexity of $O(|V|^2)$ and $O(|E|^2)$, respectively.

With the recent advances of deep learning, deep convolutional neural networks (CNN) have achieved great success, especially in the field of pattern recognition and image analysis.[8,18] Taking into account that the final output of graph visualization is an image rendered in the 2D plane, we are motivated by one crucial question: *could we evaluate the readability of graph layouts by directly using the layout images?*

In this paper, instead of using the coordinates of nodes and edges for graph layout evaluation, we propose a novel deep learning-based approach to assess the readability of graph layouts based on the layout images. It consists of a training stage and a prediction stage. In the training stage, we first generate network data with a different number of nodes using the state-of-the-art graph generation technique[10] and random walk based sampling of real graphs. Then, we take the popular force-directed graph layout algorithm as a benchmark technique to visualize all the generated networks and store both the layout image and the coordinates of nodes and edges. We further follow the traditional methods to compute the readability values by using the node and edge coordinates. These readability values, along with the graph layout images, are used for training the deep learning model. After the training is done, the deep learning model can be applied to predict the readability metrics of the input layout images. We quantitatively compared our approach with the traditional methods by using both a synthetic dataset and public real graph datasets, where both the time cost and accuracy are evaluated. Possible usage scenarios are also discussed to demonstrate the potential of the proposed approach further. The primary contributions of this work include:

- A novel deep learning based approach to evaluate the readability of graph layouts by directly using the graph images.
- A detailed quantitative comparison between the proposed approach and the traditional methods based on both synthetic and real graph datasets, which demonstrates the effectiveness and efficiency of the proposed deep learning approach.
- A use study on the potential usage scenarios that provide support for the usefulness of the proposed approach.

## RELATED WORK

The related work of this paper can be divided into two categories: deep learning and graph drawing readability metrics. We mainly discuss the representative work due to the limited space.

### A. Deep Learning

Deep learning has achieved great success in recent years and shown significantly better performance in pattern recognition and image recognition tasks than many previous conventional methods. Inspired by the notable success of CNNs, especially their application in image classification and recognition, we propose a novel CNN architecture to learn the latent features in the graph layout images and further predict their readability values. Since graph layout images usually show more apparent and relatively simpler feature structures when compared with natural images, a very deep network for identifying hierarchical features may not be necessary. Therefore, we followed the vital design principles in VGG,[18] as VGG achieves excellent results in learning the features of natural images with a relatively shallow neural network. We also adopted some other classical CNN designs such as the normalization layer introduced in AlexNet,[8] the commonly-used ReLu and Elu activation functions. On the other hand, due to the internal complexity and nonlinear structure of deep neural networks, many visualization researchers have proposed different visual analytics techniques to help people better interpret various deep neural networks, such as CNN,[12,16,20] Recurrent Neural Networks (RNN),[14] deep generative models.[11] A more comprehensive summary of these studies can be found in some recent surveys.[1,6] Inspired by them, we also visualize the basic information our CNN model to support its effectiveness.

## B. Graph Drawing Readability

The readability of a graph drawing often refers to how well a graph drawing conforms to the desired aesthetic criteria (e.g., minimizing the edge crossings and reducing the occlusions between nodes). Prior work has evaluated the readability of graph drawings through either qualitative human-centered evaluations or quantitative readability metrics.[19] Human-centered evaluations usually assess the readability of graph layouts by conducting some user studies and asking participants to finish some graph drawing related tasks,[13] with eye tracking techniques or task-oriented analysis used to indicate the readability of the graph layouts. On the other hand, readability metrics define quantitative measurements to indicate the graph readability. For example, Purchase et al.[17] investigated seven common aesthetic criteria in graph drawing and accordingly defined formal global readability metrics. Dunne et al.[3] also introduced both local and global readability metrics for the common aesthetic criteria including edge crossing, node occlusion, edge tunnel and group overlap. The calculation of the readability metrics is always based on the coordinates of nodes and edges and usually is not computationally efficient.

In contrast to prior studies, this paper attempts to estimate the readability metrics directly from the graph layout images by training a CNN model. This work focuses on the estimation of global readability metrics, which is also different from a recent work[7] that targets at the aesthetic discrimination of a pair of graph layouts. Specifically, we choose ten representative readability metrics to demonstrate the effectiveness of the proposed deep-learning-based approach in evaluating the graph layout readability.

# BACKGROUND

In this section, we provide an in-depth explanation of the choice of graph drawing algorithm and readability metrics used in our work. We also introduce the notations used in our paper.

## Why Force-Directed Layouts?

The field of graph drawing and layout analysis has been studied extensively in the past decades.[9,19] To evaluate whether a CNN-based approach would be able to learn graph metrics based on images, we decided to focus on graphs created by one specific layout. This allows us to more effectively identify and deal with any issues the model might have. Gibson et al.[5] divide the field into three major categories: force-directed, dimensionality reduction, and computational improvements based methods. We chose force-directed graph layouts,[4] as force-based methods are the most ubiquitous among researchers, especially for non-specialized datasets. Force-based layouts have also gained much attention from non-visualization researchers. Many visualization libraries, e.g., *D3.js, Gephi*, also use force-directed layouts to introduce the concept of graph drawing, further enhancing its wide application in various scenarios.

## Readability Metrics

Ten types of graph readability metrics are used in our paper. Among them, three of our chosen metrics, i.e., node number ($N_n$), edge number ($E_n$), and community number ($N_c$), are straightforward counts of the visible features, which will not be discussed in this section.

**1) Node Spread ($N_{sp}$):** The dispersion of nodes in a graph layout measures the global density of the nodes in the layout. We propose a heuristic to measure the node dispersion, which evaluates the average distance of each node from the center of its cluster. A smaller value indicates denseness and a larger value indicates sparsity. Let $A_C$ be the set of all communities. Every node belongs to exactly one community in $A_C$. We denote the $x$ and $y$ values of a node $i$ as $i_x$, $i_y$ respectively. We denote the mean $x$ value and mean $y$ value of all nodes in community $c$ as $\mu_x^c$, $\mu_y^c$ respectively. Node Spread is defined as follows:

$$N_{sp} = \sum_{c \in A_c} \frac{1}{|c|} * \sum_{i \in c} \sqrt{(\mu_x^c - i_x)^2 + (\mu_y^c - i_y)^2} \quad (1)$$

**2) Node Occlusions ($N_{oc}$):** Node Occlusions ($N_{oc}$) is the sum of all instances of nodes being positioned at the same coordinates within a threshold of closeness. To compute $N_{oc}$, we iterated over all node pairs and incremented our counter if two nodes have overlapping boundaries.

**3) Edge Crossings ($E_c$, $E_{c.outside}$):** Edge crossings can make a graph look cluttered and obfuscate information about inter-node relationships, so minimizing this metric is helpful for accurate perception. The calculation and usage of edge crossing as a measure of graph readability have been discussed by Dunne et al[3]. We implemented two edge-crossing-based metrics, general edge crossings ($E_c$) and edge crossings outside communities ($E_{c.outside}$). The former is defined as the count of pairwise edge intersection within a graph layout, while the latter counts the overlaps occurring between communities, i.e., where the start and end nodes of an edge belong to different communities.

**4) Minimum Angle ($M_a$):** Minimum angle ($M_a$) was originally defined by Purchase[17] as the average value of the absolute deviation between the ideal minimum angle $\phi_{min}(V)$ and the minimum incident angle of each edge $\phi(v)$, where $\phi_{min}(V)$ is defined as: $\phi_{min}(V) = 360/deg(v)$. The range of this metric can be from 0 to 1, where 0 means equally spaced edges whereas 1 means that the average deviation of the edge angle is 0, i.e., they are incident on each other. Computing the minimum angle metric is quite straightforward and has a worst case complexity of $O(E)$.

$$M_a = 1 - \frac{1}{|V|} \sum_{v \in V} \frac{\phi(v) - \phi_{min}(V)}{\phi(v)} \qquad (2)$$

**5) Edge Length Variation ($M_l$):** Edge Length Variation ($M_l$) is defined as the deviation of edge length from its mean value. Studies[9] have shown that uniform edge length is a good criterion when measuring the quality of a graph layout. $M_l$ is defined as $M_l = l_a / \sqrt{|E| - 1}$, where $l_a$ is defined as the ratio of standard deviation to mean of edge lengths in a graph drawing. $l_e$ and $l_m$ are edge-length and mean edge length respectively, while $l_\mu^2$ is used to standardize the data. This metric is normalized using $\sqrt{|E| - 1}$ as the upper bound for variation.

$$l_a = \sqrt{\sum_{e \in E}(l_e - l_\mu)^2 / (|E| * l_\mu^2)} \qquad (3)$$

The computational complexity of such metric is O(|E|), as we just need to iterate through each edge in a graph to compute the metric.

**6) Group Overlap ($G_o$):** Group Overlap ($G_o$) is defined as the number of overlaps between communities in the graph.[3] This metric is used to evaluate the clarity of group membership within a graph, as the higher the value of $G_o$, the more visually unclear group membership becomes. We calculate this metric using a modified version of the algorithm proposed by Dunne et al.[3] The modified algorithm runs in $O(gnlog(n))$, where $g$ and $n$ are the number of groups and points in the graph respectively.

# CNN-BASED APPROACH

We propose a CNN-based approach that learns the relationship between a graph layout image and its readability metrics. Such a supervised learning approach usually requires a large benchmark dataset for training. However, no benchmark dataset exists for our particular project, so we generated 193,500 graph layout images and calculated their readability metrics to obtain the ground-truth values. While this is a time-consuming task, it has to be run only once in the pre-processing phase. In this section, we will introduce the method of training dataset generation and the CNN Model architecture.

## Training Dataset Generation

In this section, we explain the dataset generation process, from generating the adjacency matrices to labeling the graph layouts.

**1) Generation of adjacency matrices:** Generating synthetic graphs is a well-studied area of research with many algorithms proposed, each focusing on optimizing a set of specific network features. In this paper, we use the method proposed by Lancichinetti et al.,[10] which focuses on creating benchmark graphs with complex community structures and can generate more realistic synthetic graphs. Also, this algorithm allows end users to provide specifications for the graph, such as the Number of nodes, number of communities, etc. Therefore, we adopted this algorithm for our adjacency matrix generation. When creating the edge lists using the algorithm mentioned above, we had to decide on input parameters. After experimenting with multiple input parameters, we finally set the number of nodes to be a uniformly distributed number from 50 to 600, the number of communities to be a uniformly distributed number from 2 to 128, and the average node degree was empirically set to a minimum of 3 edges. We chose the community-overlap parameters to be a normally distributed number in the range of 0.1 to 0.5. Using these parameters, we generated around 16,500 edge lists. Apart from generating synthetic graphs, we also sub-sampled many large real networks to collect more real networks. These real networks, e.g., *Facebook graph, Google-cfinder, Zachary's Karate graph, Dolphin social network, Les Miserable network,* are mainly downloaded from University of Koblenz networks repository[*]. We used a random walk based method to sample these graphs, where the size of the walks, the maximum number of nodes in the sampled graphs, and the nodes to be walked were randomly chosen within the ranges defined for the synthetic data generation process. Unlike synthetic graph generation, where the algorithm can assign a community label for each node, real graph generation uses the method introduced by Clauset et al.[2] to assign community labels. Using this sampling method, we generated 5,000 adjacency matrices.

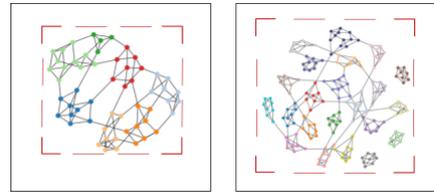

Figure 1. Cropping graphs out from images to remove additional white regions from the input graph image.

**2) Rendering of Force-Directed Graph Layouts:** In our implementation of the force directed graph layout algorithm, we choose gravity from {0.3, 0.4, 0.5} and charge from {−300, −400, −500} and generate nine layout configurations. Given these configurations, we then created nine different graph layouts for every graph. These graph layouts were then drawn on a canvas with a size of 1620×1350. The radius of each node is set to 8 pixels, and the width of every link is set as 0.5 pixels. We cropped the main rendering region of graph layouts and removed surrounding white space, focusing on the region of interest (as shown in Figure 1). Finally, all the graph layout images are resized to a size of 325×260. This resulted in 193,500 graph layouts including both real and synthetic datasets.

**3) Labelling of Graph Layouts:** When rendering the graph layouts in the previous step, we also stored the coordinates of nodes and edges. This makes it possible for us to compute readability metrics using the methods described in the Background section and save them as labels for that layout. Some example images of the dataset are shown in Figure 2, where the ten readability metrics introduced in the Background section are calculated for each layout, with the scores shown below. These metric scores are used in the training process.

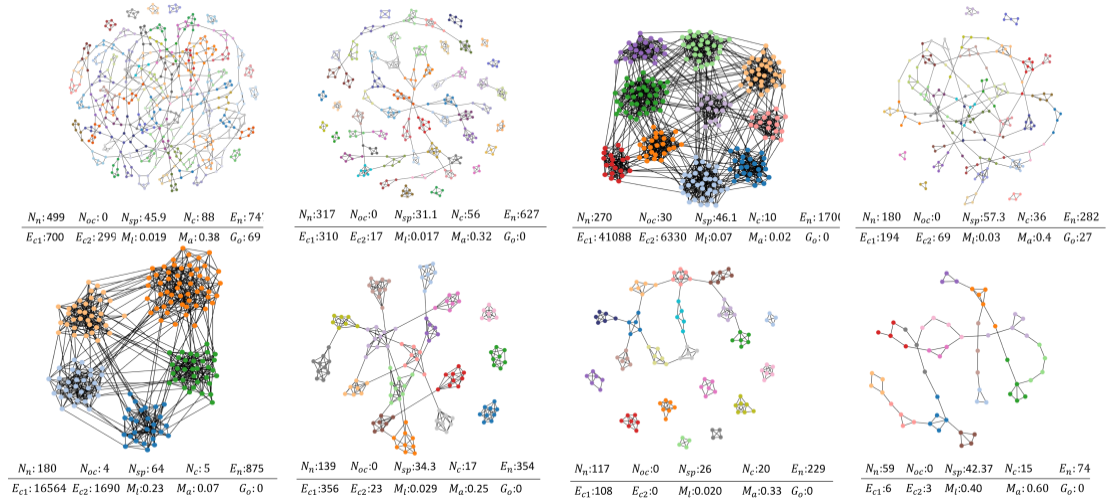

Figure 2. Eight examples of graph drawings from our training dataset, and each drawing is augmented with labels. All the image labels are defined in the Background section and the images are sorted by $N_n$.

## Model Architecture

CNNs are a type of deep learning neural networks that have recently been widely used for tackling image-related learning tasks. A large number of CNN architectures have been proposed and various CNN layers are also discussed.[8,18] Compared to standard feed-forward neural networks, CNNs have similarly-sized layers but with much fewer connections and parameters. Therefore, they are easier to train, while their theoretically-best performance is likely to be only slightly worse.[8] Our models' architecture is inspired by AlexNet and VGG.[18]

As shown in Figure 3, the input of our CNN model is a three channel image with a size of 325 × 260. Proposed CNN model has six convolutional layers, where the first two layers have 32 filters and the last four layers have 64 filters each. Between the first two layers, we add local response normalization (LRN) to normalize the input and thus improve the training. When designing a convolutional layer, it was observed that a stack of three 3×3 filters has the same effective receptive field as one 7×7 convolutional layer, as introduced in the VGG.[18] The stacked 3×3 filters allow the network to learn the hidden features more accurately, make the network deeper, gain more non-linearities and use fewer parameters. Therefore, stacked 3×3 filters are mainly used in our convolutional layers. Both the stride for the convolutional layer and the spatial padding are all set as 1 pixel in our CNN architecture.

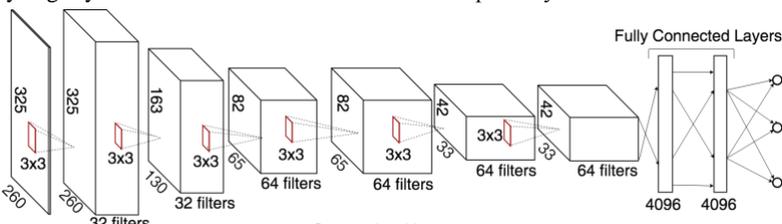

Figure 3. Our proposed CNN design, with six convolutional layers, two fully connected layers and three output neurons.

[*] http://konect.uni-koblenz.de/

After each convolutional layer, we perform Max-pooling over a 2×2 pixel window with stride two. Finally, we have 2 fully connected layers with a size of 4096, followed by individual fully connected neurons, where each neuron corresponds to a specific metric. For example, in the node feature learning model, we added four neurons after the fully connected layer to predict the four node based metrics. When CNN is applied in a classification problem, a fully connected layer with a SoftMax activation function is usually used to generate probabilities for each class. By changing this activation function, however, it is possible to use a very similar model for a regression problem, as demonstrated in our approach. In the last layer, we replace the SoftMax activation function with the ReLu function. This changes the behavior of the network, as previously the SoftMax would map the output to the range of 0 to 1, whereas the ReLu now allows the network to output positive values in a real domain (0 to ∞), targeting at a regression problem (i.e., predicting the graph readability metric values). Using the above architecture, we trained three models for the three different groups of metrics:

- Node feature learning model: 4 output neurons predict $N_n$, $N_{oc}$, $N_c$ and $N_{sp}$.
- Edge feature learning model: 3 output neurons predict $E_n$, $E_c$, and $E_{c.outside}$.
- Global metric learning model: 3 output neurons predict $M_a$, $M_l$ and $G_o$, respectively.

## Implementation Details

During the training process, we use a quadratic loss function where the optimizer tries to minimize the mean of the square of differences between the actual and predicted metric values. For the training settings, we empirically set the mini batch size as 50 images, and the learning rate as $10^{-4}$. Adam Optimizer is the optimizer used in our experiments. The proposed CNN-model was implemented using TensorFlow. The testing and evaluation of the model were run on two servers each with 4 Nvidia™ graphics cards with the latest Tesla architecture (GTX 1080), and 128GB RAM with Intel Xeon processors (CPUE5-2650 V4). The training process itself took four days to finish 20 epochs, where one epoch is one iteration over every data item in the dataset.

## EVALUATION

To demonstrate the effectiveness of the proposed CNN model, we comprehensively evaluated it from different perspectives. We first conducted K-fold validation on synthetic graph layout datasets. Then, we tested if adding real graphs into the training dataset will improve the testing performance. We also visualized the convolutional layers and compared the computational speed to evaluate the effectiveness of the proposed approach further.

## K-fold Validation on Synthetic Dataset

Our dataset was randomly partitioned into three parts, where two were combined into a training dataset, and the third was used as a testing dataset. We then rotated the three datasets around so that every dataset was chosen as the testing dataset exactly once. We observed that the CNN model quickly converged and the loss decreased significantly with more iterations. Once the three testing datasets have been executed, we evaluated the average absolute loss using confidence intervals (CI). For each run in the cross-validation phase, the proposed model achieved an R2 score of over 0.85, indicating that the model is able to learn the underlying pattern in the data.

Table 1. The 95% confidence interval of the mean value of the metrics across all training sets. The average normalized value of the real input value is shown next to the metric.

| Training Set | $N_n$: 44.91 | | $N_{oc}$: 3.9 | | $N_c$: 30.22 | | $N_{sp}$: 23.04 | | $E_n$: 8.07 | | $E_c$: 124.98 | | $E_{co}$: 22.24 | | $M_a$: 26.11 | | $M_l$: 2.42 | | $G_o$: 3.55 | |
|---|---|---|---|---|---|---|---|---|---|---|---|---|---|---|---|---|---|---|---|---|
| | Low | High | Low | High | Low | High | Low | High | Low | High | Low | High | Low | High | Low | High | Low | High | Low | High |
| 1 + 2 | 0.87 | 0.89 | 0.64 | 0.67 | 1.25 | 1.27 | 1.06 | 1.08 | 1.24 | 1.26 | 8.23 | 8.68 | 2.23 | 2.34 | 1.42 | 1.45 | 0.18 | 0.18 | 1.37 | 1.43 |
| 1 + 3 | 0.67 | 0.68 | 0.78 | 0.81 | 1.37 | 1.39 | 1.15 | 1.18 | 0.85 | 0.86 | 6.56 | 6.90 | 1.64 | 1.72 | 1.39 | 1.42 | 0.23 | 0.23 | 1.35 | 1.40 |
| 2 + 3 | 0.58 | 0.59 | 0.73 | 0.76 | 1.15 | 1.17 | 1.13 | 1.15 | 1.08 | 1.10 | 7.66 | 8.09 | 2.09 | 2.19 | 1.42 | 1.45 | 0.18 | 0.18 | 1.38 | 1.43 |

*1) Evaluation of Absolute Loss:* For each of these models, we computed the average absolute loss along with a 95% CI. We then compared the CI to the average values for each metric to convert the CI from absolute loss to percentage loss, giving us an estimate of the accuracy of the model. To calculate the confidence interval, we first let $X = (x_1, x_2, ... , x_N)$ be the set of all the input values (ground truth) and $Y = (y_1, y_2, ... , y_n)$ be the set of all the predicted values for metric $i$ by the model. Then $Z = (|x_1 - y_1|, |x_2 - y_2|, ...., |x_n - y_n|)$ is the set of absolute differences between $X$ and $Y$. Given these definitions, the 95% *CI* for metric $i$ is given as below:

$$CI_i = [\mu - \frac{1.96\,\sigma}{\sqrt{N}}, \mu + \frac{1.96\,\sigma}{\sqrt{N}}] \quad (4)$$

Where $\mu$ is the mean and $\sigma$ is the standard deviation of Z. It is important to note that our confidence intervals are based on the absolute loss, not the percentage loss. We decided to take this approach because, for some of our metrics, the inputted value was 0. This would create issues if we used percentage losses. However, having a CI of the absolute loss isn't useful for comparison across metrics, so we calculated the high bound of the CI as a percentage of the average value for that metric, called P, as shown below:

$$P = \frac{CI_{High}}{\frac{1}{N} * \sum X_n} \quad (5)$$

This allows us to compare the model's performance across metrics. Table I presents the absolute CI and the average input value for the metrics. Figure 4 shows the percentage errors computed using our CI. We would like to note that our results for Go are not included in Figure 4 as it would distort it, but the average percentage error was 39%, a marked anomaly. Our results show that, barring $N_{oc}$ and $G_o$, the average percentage error ranges from 5 to 15%, meaning that with 95% confidence our results are 85+% accurate. We see this as an extremely strong result, especially considering the novelty of this approach to the problem. The results for $N_{oc}$ and $G_o$, while relatively weak, reflect the inherent difficulty that even an expert would have when evaluating these metrics. This difficulty is because, even to the naked eye, nodes on top of each other look identical and cannot be distinguished. As these two metrics aim to measure such phenomena, it is understandable that the error is high.

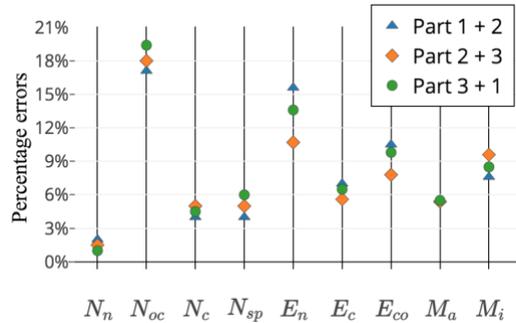

However, the model on which we performed the K-fold analysis was subsequently augmented with two more convolutional layers. This was done to speed up the training process by decreasing the number of connections between the fully connected and convolutional layers. To ensure that our changes had not adversely impacted the error scores, we re-ran the training process on Parts 1+2 and tested on Part 3 and observed that the error had dropped. This indicates that this change is not increasing the error rate of our model.

Figure 4. Percentage errors over all sets in our K-fold evaluation for all readability metrics on synthetically generated data.

## Testing on Publicly Available Real Graphs

Since a large number of real graphs are currently not available, we created a validation dataset with a combination of both common real graphs and sampling large real graphs, which was discussed in the CNN-based Approach section. It should be noted that certain hairball like graphs were removed from the dataset, as the model is expected to fail on such graphs, given their visual complexity. For our dataset, we found any graph with more than 5000 edge crossing would result in a hairball like structure. As these graphs do not come with a ground-truth value for community structure, we computed it using the fast-greedy algorithm,[2] a well-known community computation measure. This allowed us to calculate $G_o$, $N_{sp}$ and $E_{c.outside}$ as these metrics are highly dependent on community information. We then utilized traditional methods to compute other metrics for these graph images.

Once all the layouts in the training set have been generated, we then test our models' performance on our validation dataset. Two versions of the model were tested; one version was trained on only synthetic data while the other version was trained on a combination of synthetic and real data. The first version performed well on the simple metrics but failed on more complicated metrics such as group overlaps and node spread. Our second version performed markedly better. As reported in Table 2, Row P.E.2 shows the percentage errors for experiment case 1 where we trained our model only on synthetic data (M.1) and Row P.E.1, shows the results when we trained on a combination of both synthetic and real data (M.2).

Table 2. Percentage errors for the real dataset. P.E.1 is the percentage errors when the model is trained on a combination of synthetic datasets and real datasets. P.E.2 is the percentage error when the model is trained on synthetic dataset only.

| Metric | $N_n$ | $N_{oc}$ | $N_c$ | $N_{sp}$ | $E_n$ | $E_c$ | $E_{co}$ | $M_a$ | $M_l$ | $G_o$ |
|---|---|---|---|---|---|---|---|---|---|---|
| P.E.1% | 8.30% | 22.20% | 22% | 12.50% | 39.07% | 55.88% | 75.20% | 23.22% | 40.22% | 30.77% |
| P.E.2% | 6.01% | 11.1% | 10.2% | 12.39% | 5.5% | 16.3% | 17.40% | 17.22% | 11.2% | 18.83% |

We can see both the models performed well when computing simple graph metrics, overall *M.2* performs much better, especially on edge-based metrics. The relatively high error in the $E_{C.outside}$ and $E_{c.c}$ metrics is mainly due to very complicated graph structure in specific cases where even humans cannot correctly evaluate such graph. These metrics are related to the underlying community structure, for which there is no ground-truth value. We believe this lack of ground-truth contributes to the higher error. We also noted during testing that the inclusion of graphs with highly overlapping community structures and strongly connected node structures had a considerable impact on the overall performance of the model.

### Visualizing the Convolutional Layers

As part of our evaluation of the CNN model, we visualized the neuron activation of all the convolutional layers. This would allow us to see if the layers are learning features of the input image. We show the results of two input graph layout images here, where one network was synthetically created and well visualized, while the other was sampled from a real graph and has more visual clutter than the first graph layout. By doing this, we were able to see if the model was adequately handling these two types of inputs. For analysis, we mainly focused on the first two convolutional layers, as the following layers were highly abstract representations of the models learning, and not interpretable by the human eye. From Figure 5, we can see that the first two layers do exhibit clear signs of feature learning. In Figure 5, we can observe (a), (b) and (e) are learning simple features of the image (boundaries, edges, and circles), whereas we can see in (c) it is learning about node groups. In the Conv1, we can see CNN starts to learn more complex features like network structure (g) and dense edge areas (d). For both the kinds of input images, the proposed model is able to detect image features with a high level of detail and accuracy, providing a qualitative affirmation of our models' ability to learn the readability features of graph layouts.

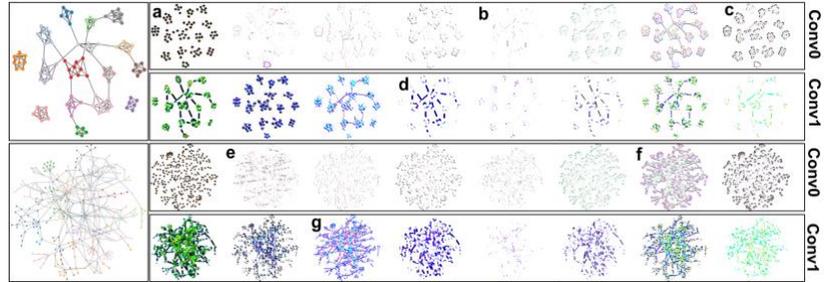

Figure 5. Two sample input images and the corresponding outputs of the first two convolutional layers.

### Computational Speed Comparisons

The computational speed is, theoretically, one of the main advantages of using a learning-based model compared with computing metrics using traditional methods. To quantitatively measure the speedup, we chose a random set of 50,000 images and timed how long the model takes to predict all the metrics vs. how long traditional methods take. We show the spread of these times per input in Table 3 with respect to the number of nodes. From the results, we can see that our model vastly outstrips the traditional methods, with a speedup factor larger than 10. Note that we do not use GPU to speed up the calculations of the proposed CNN model and all the calculations are run on the same CPU. This result reflects the fact that the running time of our model is invariant to the graph size, as the size of the input image is always the same. The computational speedup, combined with the previous evaluation results, demonstrates the strong advantages of our model over the traditional readability evaluation methods. One possible explanation for the high speed of our method is that it is solely based on the input images and is invariant with respect to the graph size.

Table 3. Time comparison for computing layout labels with the CNN model and traditional methods, where T is the computational time for traditional methods and C is the computational time for the CNN model, both in seconds.

| Avg. $N_n$ | $N_n$ <100 | | 100<=$N_n$<200 | | 200<=$N_n$<300 | | 300<=$N_n$<400 | | 400<=$N_n$<500 | | 500<=$N_n$<600 | |
|---|---|---|---|---|---|---|---|---|---|---|---|---|
| T/C in sec | T | C | T | C | T | C | T | C | T | C | T | C |
| Node Metrics | 0.053 | **0.023** | 0.218 | **0.030** | 0.489 | **0.023** | 0.995 | **0.022** | 1.595 | **0.27** | 2.734 | **0.021** |
| Edge Metrics | 0.598 | **0.011** | 3.095 | **0.014** | 5.254 | **0.014** | 10.673 | **0.014** | 16.050 | **0.011** | 11.28 | **0.011** |
| Others Metrics | 0.052 | **0.010** | 0.175 | **0.014** | 0.368 | **0.011** | 0.698 | **0.012** | 1.123 | **0.011** | 0.529 | **0.011** |

## USAGE SCENARIOS

In this section, we describe a usage scenario to demonstrate how the proposed method can help users to easily evaluate graph layouts and quickly select the graph layouts with the preferred readability. We further discuss how the proposed technique could be applied in other applications.

## Scenario 1: Selecting the Desired Graph Layouts

Assume Andy, a software engineer at a search engine firm, is attempting to index the graph images they have. Such an index would allow users to search for "graphs with lots of edge crossings", "3 community graph", etc., which increases the ability of users to access specific graph layout images through the engine. As of now, the firm has millions of graph images that need to be indexed for such kind of searching, and the firm's crawlers will constantly be adding any new graph images they find to this database. To do this index, it is important for Andy to know what is the underlying readability of every graph drawing layout image so that they can be properly cataloged. But with traditional methods, the computational complexity is too high and would take an excessive amount of time to go through all such images, even if we disregard the constant stream of new graph images coming in. Also, certain metrics would not be possible to calculate without knowing the adjacent matrix of the graph in the image. However, with our approach, there is a computational speedup factor of at least 100x, and our model can be trained to predict essentially any graph metric.

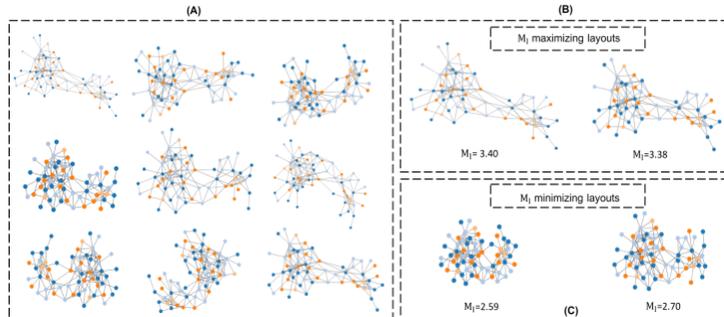

Figure 6. Given the graph layouts in (A), the proposed model can predict their $M_l$. The graph layouts with top-2 maximum and minimum $M_l$ are present in (B) & (C), respectively.

Andy isn't the only one who has issues in this area. Dave, a web designer, faces a challenging problem at work. His project requires an image of a graph that shows a lot of communities, and he has access to a bank of thousands of images of different graphs. Such banks can be found in design firms as they often need to use esoteric images in their work. However, manually going through this bank to find the specific graphs would be a highly laborious task, and traditional methods would take a significant amount of time to process all these images as well. Our approach can be applied to such situations, as our model would allow for all the images to be evaluated and the graphs with the highest community metric to be returned. For example, if his aim was to minimize $M_l$ then the graph layout best fitting this requirement would be returned, as shown in Figure 6. The above scenarios highlight the benefits of using our proposed model to compute graph metrics for images.

## Other Usage Scenarios

There are multiple scenarios in which our proposed technique can be beneficial. One such scenario is image searching and indexing. A search engine, such as Google, will have millions of graph images. Creating a metric-based index of these images is computationally expensive with traditional methods; our method allows for more efficient metric-based indexing, enabling users to search using metric related keywords. With regards to applications in visualization, there are streaming-based scenarios where our technique provides value. If a system is visualizing a high bandwidth data stream, traditional methods will lag behind the stream. This means that users will be presented with outdated metric values. Our technique would allow the displayed metric values to be essentially real-time, creating a smoother user experience and, in critical scenarios, reducing the response time of the users to issues.

## DISCUSSIONS

Our quantitative evaluation and usage scenarios have demonstrated the effectiveness and usefulness of the proposed CNN-based approach in evaluating the readability of graph layouts. It is much faster and can predict the readability metrics with high accuracy. But there are still several issues that need to be further elucidated.

**Training datasets:** Currently, the training datasets are a combination of synthetic datasets, where the graph data is first generated from an automated graph generation algorithm and sampling of real graphs. The ground truth readability values are assigned by using traditional methods. It would be better if all the training graphs are real graphs and the readability values are manually labeled by humans, which can reflect the actual readability appreciated by a human. However, such large datasets are still not available. Therefore, we choose to train our CNN model on this combination of synthetic and sampled real graphs.

**Evaluating edge-related readability metrics:** As shown in the Evaluation section, the proposed CNN model has a relatively better accuracy in evaluating node-related readability metrics than edge-related readability metrics. One possible explanation for this would be that it is difficult for the CNN model to learn the details of edges and edge crossings when the dense nodes occlude the edges or the edges themselves are too dense or highly overlapped to extract any useful information through images. Also, the proposed method only focuses on the general node-link diagrams and does not take into account the cases with edge bundling.

**Time cost of the approach:** One might argue that the training time makes the time-efficiency comparison unfair, as discussed in the evaluation section. While it is true that training the model takes a significant amount of time, once the model has been trained it can be readily applied by other users with no extra processing.

**Limit on max number of nodes:** As a proof of concept, this paper mainly tests the graph drawing images with less than 600 nodes, since the image size we can feed to a CNN model is limited by a number of factors including the memory, the size of CNN, as the number of parameters increases when we are training on a larger image. Furthermore, increasing the number of nodes by more than 600 on current image size, we would not be able to draw it without significant information loss.

**Potential improvement directions for the proposed approach:** Despite the good performance of the proposed deep learning model for graph readability evaluation, we believe there are still several directions that could be explored further, to enhance the performance of the proposed method. First, as mentioned above, when a large real graph dataset is available, It could be utilized to train the CNN model, giving better predictions. Also, the current synthetic training dataset is mainly rendered by using the force-directed graph layout algorithms. Many other layout algorithms can also be used to generate more diverse layout images to improve the generality of the proposed approach further. Moreover, as a common practice in deep-learning,[8,18] model hyper-parameters can be further tuned to improve the overall performance.

## CONCLUSION AND FUTURE WORK

Traditional quantitative methods for evaluating the graph readability are based on the coordinates of nodes and edges in the graph layout. Such methods require extra space to store the coordinates and are usually not computationally efficient, especially when the graph size increases. In this paper, we propose a CNN-based model to evaluate the readability of graph layouts, where the readability metrics of a graph layout are evaluated by using the graph layout result itself (i.e., the graph drawing image). We generated the graph layout dataset by using the prior representative algorithms and label these layout images with readability values evaluated through traditional methods as the ground truth. Then, these graph layout images, as well as the assigned readability metric values, are used to train the proposed CNN model. Three CNN models are trained for the three different classes of readability metrics. In the testing stage, the trained CNN model directly predicted the readability values, given a graph layout image. Our experiment showed that our approach is more than ten times faster than the traditional readability calculation methods and has a consistent, efficient performance regardless of graph size. Also, it achieved a high level of accuracy, especially for node-related readability metrics. The usage scenarios further provide support for the usefulness of the proposed approach.

In future work, we plan to further train the proposed CNN model on manually-labeled real graph datasets instead of synthetic graph datasets and further evaluate the effectiveness of our approach. Also, we would like to extend the proposed approach to the evaluation of other visualization designs such as scatterplots, parallel coordinates, bar chart, and parallel coordinates.

## ABOUT THE AUTHORS

**Hammad Haleem** is a strategist at Goldman Sachs. He received his M.Phil and M.Sc degrees from the Hong Kong University of Science and Technology in Computer Science, and B.Tech from Jamia Millia Islamia, New Delhi. Hammad's research interests include computer graphics, visual analytics, and machine learning. This work was done as part of his M.Phil research thesis. Contact him at hammadhaleem@gmail.com.

**Yong Wang** is currently a post-doctoral fellow in the Department of Computer Science and Engineering at the Hong Kong University of Science and Technology (HKUST). He received his Ph.D. degree in Computer Science and Engineering from HKUST in 2018. His research interests include graph visualization, visual analytics, and machine learning. He is the corresponding author of this paper. Contact him at ywangct@cse.ust.hk.

**Abishek Puri** is an M.Phil student in Computer Science at the Hong Kong University of Science and Technology. His research focus is on building predictive models for specific datasets and designing visualizations that allow users to explore the results of the models fully. Contact him at apuri@connect.ust.hk.

**Sahil Wadhwa** Is an analyst working in Financial Modelling Group in Blackrock, India in the field of Natural language processing. He completed his undergraduate in computer engineering from Jamia Millia Islamia, New Delhi, India. Contact him at sahil24wadhwa@gmail.com.

**Huamin Qu i**s a full professor in the Department of Computer Science and Engineering (CSE) at the Hong Kong University of Science and Technology. He also serves as the coordinator of the newly founded Human-Computer Interaction (HCI) group at the CSE department.  He obtained a BS in Mathematics from Xi'an Jiaotong University, China, an MS and a Ph.D. (2004) in Computer Science from the Stony Brook University. Contact him at huamin@cse.ust.hk.